\pdfoutput=1
\documentclass[journal]{IEEEtran}
\IEEEoverridecommandlockouts
\usepackage{color}
\usepackage{subfigure}
\usepackage[ruled,linesnumbered]{algorithm2e}

\usepackage{graphicx}
\usepackage{amssymb}
\usepackage{amsmath}
\usepackage[numbers,sort&compress]{natbib}

\usepackage{booktabs}
\usepackage{multirow}

\usepackage{float}     

\usepackage{multirow}
\usepackage[hidelinks]{hyperref}

\begin{document}

\title{A Graph Foundation Model for Wireless Resource Allocation}

\author{Yucheng Sheng,~\IEEEmembership{Graduate Student Member,~IEEE,}
        Jiacheng Wang,~\IEEEmembership{Graduate Student Member,~IEEE,}
        \\Le Liang,~\IEEEmembership{Member,~IEEE,}
        Hao Ye,~\IEEEmembership{Member,~IEEE,}
        and Shi Jin,~\IEEEmembership{Fellow,~IEEE}
\thanks{Yucheng Sheng, Jiacheng Wang, Le Liang, and Shi Jin are with the National Mobile Communications Research Laboratory, Southeast University, Nanjing 210096, China (e-mail: shengyucheng@seu.edu.cn; wangjiacheng@seu.edu.cn; lliang@seu.edu.cn; jinshi@seu.edu.cn).}
\thanks{Hao Ye is with the Department of Electrical and Computer Engineering, University of California, Santa Cruz, CA 95064, USA (e-mail: yehao@ucsc.edu).}
}

\maketitle

\begin{abstract}
The aggressive densification of modern wireless networks necessitates judicious resource allocation to mitigate severe mutual interference. However, classical iterative algorithms remain computationally prohibitive for real-time applications requiring rapid responsiveness. While recent deep learning-based methods show promise, they typically function as task-specific solvers lacking the flexibility to adapt to different objectives and scenarios without expensive retraining. To address these limitations, we propose a graph foundation model for resource allocation (GFM-RA) based on a pre-training and fine-tuning paradigm to extract unified representations, thereby enabling rapid adaptation to different objectives and scenarios. Specifically, we introduce an interference-aware Transformer architecture with a bias projector that injects interference topologies into global attention mechanisms. Furthermore, we develop a hybrid self-supervised pre-training strategy that synergizes masked edge prediction with negative-free Teacher-Student contrastive learning, enabling the model to capture transferable structural representations from massive unlabeled datasets. Extensive experiments demonstrate that the proposed framework achieves state-of-the-art performance and scales effectively with increased model capacity. Crucially, leveraging its unified representations, the foundation model exhibits exceptional sample efficiency, enabling robust few-shot adaptation to diverse and unsupervised downstream objectives in out-of-distribution (OOD) scenarios. These results demonstrate the promise of pre-trained foundation models for adaptable wireless resource allocation and provide a strong foundation for future research on generalizable learning-based wireless optimization.
\end{abstract}




\IEEEpeerreviewmaketitle

\section{Introduction}

The emergence of sixth-generation (6G) wireless ecosystems is predicated on the imperative to support ubiquitous and massive connectivity \cite{qin2024ai}. A key characteristic of this evolution is the aggressive densification of wireless networks, which aims to improve spectral efficiency through extensive spatial reuse of spectrum resources. However, recent studies have revealed several inherent issues of such hyper-connected environments, particularly the severe mutual interference caused by the spatial reuse of spectrum \cite{dai2024survey}. Consequently, it is of paramount importance to judiciously allocate these wireless resources to mitigate interference in such dense networks. 

Mathematically, resource allocation over interference channels is generally a non-convex and NP-hard problem. While classical iterative algorithms, such as weighted minimum mean squared error (WMMSE) \cite{christensen2008wmmse} and fractional programming (FPLinQ) \cite{shen2018fp1, shen2018fp2}, can converge to stationary points, they are hindered by high computational complexity and slow convergence rates, making them difficult to scale to large networks. 
To alleviate computational overhead, early deep learning (DL) approaches utilize multi-layer perceptrons (MLPs) \cite{sun2018learning}, convolutional neural networks \cite{cui2019spatial}, and reinforcement learning \cite{liang2019spectrum}. Despite accelerating inference, they fail to capture the inherent permutation symmetries in wireless interference scenarios. This limitation motivates the adoption of graph neural networks (GNNs) \cite{shen2020graph, shen2022graph, naderializadeh2022state}, which efficiently mimic iterative solvers by explicitly modeling graph topologies. Nevertheless, the performance of rigidly trained GNNs often degrades in non-stationary environments with varying network topologies and fluctuating channel state information (CSI). To achieve robust generalization, recent paradigms integrate meta-learning with GNNs \cite{zhao2024meta, nikoloska2023modular, hou2023meta}, enabling rapid adaptation to dynamic network conditions using only a few samples.

Despite these advancements, a fundamental issue exists in these learning-based frameworks as they predominantly function as task-specific solvers, i.e., to optimize a single, specific objective. However, practical wireless networks often need to accommodate a wide variety of different design objectives, such as sum rate maximization, proportional fairness (PF), quality-of-service (QoS)-aware optimization, etc. Whenever the design goal changes, these learning methods, trained to optimize one specific objective, inevitably need expensive retraining, often from scratch. This limitation severely hinders the widespread adoption of learning-based wireless resource allocation methods in practice.
To address these issues, recent research has begun exploring unified large-scale architectures \cite{liang2026large}, drawing inspiration from the success of foundation models in natural language processing \cite{devlin2019bert, brown2020language} and computer vision \cite{dosovitskiy2021image}. These models are typically pre-trained on large-scale data to learn transferable representations and then adapted to downstream tasks. In wireless communications, existing efforts have primarily focused on physical-layer signal processing. Specifically, frameworks such as the LWM \cite{alikhani2024lwm}, WirelessGPT \cite{yang2025wirelessgpt}, and WiFo-2 \cite{liu2025wifo2} utilize self-supervised learning and reconstructive autoencoders to extract generalized representations from massive channel datasets. While these methodologies have proven effective for physical-layer tasks, including beam prediction and channel estimation, they fundamentally lack the mechanism to capture the topological dependencies and the underlying interference patterns within multi-user networks, which is crucial for effective resource allocation. Consequently, extending the foundation model philosophy to manage network-level interference topology remains a critical, unresolved challenge.

To bridge this gap, we propose a graph foundation model tailored to wireless resource allocation. By pre-training on massive unlabeled data, this foundation model captures a highly transferable structural representations, enabling sample-efficient adaptation to heterogeneous downstream resource allocation tasks. Realizing this paradigm requires a backbone architecture that is both expressive and scalable. While MPNNs are widely used for wireless graph learning, their capacity is fundamentally limited by over-smoothing \cite{li2018deeper, chen2020measuring}, which causes node representations to converge and hinders knowledge accumulation during large-scale pre-training. Transformer architectures, by contrast, have demonstrated strong scalability and have become a natural choice for large-scale representation learning. Moreover, their global self-attention mechanism is well suited to modeling the dense interactions that arise in highly connected interference graphs. However, standard Transformers do not explicitly encode graph topology and therefore cannot directly exploit the relational structure of wireless networks. Motivated by structural modeling advances such as Graphormer \cite{ying2021do} and AlphaFold2 \cite{jumper2021highly}, we introduce a bias projector that maps edge features in an interference graph into the attention mechanism. By injecting this physical information as bias terms into the attention scores, we endow the Transformer with the ability to perform physically aware global reasoning, combining the scalability of Transformer with the topological precision required for wireless networks.

With a scalable and interference-aware backbone established, the next challenge lies in designing self-supervised objectives that can inject transferable physical priors into the model. Existing graph pre-training strategies generally fall into two paradigms: generative and contrastive. Existing generative frameworks, such as generative pre-training of graph neural networks (GPT-GNN) \cite{hu2020gptgnn} and masked graph autoencoders (GraphMAE) \cite{hou2022graphmae}, primarily focus on the reconstruction of node features or the prediction of discrete link existence \cite{hu2020strategies}. Such objectives are not fully aligned with wireless interference graphs, where the most informative relational structure is carried not only by node states but also by continuous edge features that represent interference strength. To better capture this structure, we introduce a masked edge prediction objective in which the model reconstructs missing continuous interference values from the surrounding graph context. This task encourages the backbone to model the spatial correlations and relational patterns that govern interference across users. 

However, generative reconstruction alone is insufficient. During pre-training, the model is exposed to masked graphs, whereas downstream deployment may operate on complete graphs, creating a mismatch between the training and inference views. Solely relying on edge prediction may lead to representations that are over-dependent on the masking pattern. To improve robustness to this discrepancy, we complement masked edge prediction with a contrastive consistency enforcement. While classical graph contrastive methods like graph contrastive learning (GraphCL) \cite{you2020graph} and deep graph infomax (DGI) \cite{velickovic2019deep} rely on negative sampling, constructing effective negative pairs in fully connected interference graphs is conceptually difficult and computationally expensive. Therefore, we propose a negative-free Teacher-Student architecture, inspired by bootstrapped representation learning (BGRL) \cite{thakoor2021bootstrapped} and simple graph contrastive learning (SimGRACE) \cite{xia2022simgrace}. By maximizing the similarity between the representations of the masked online view and the original target view, we ensure that the learned embeddings remain consistent and robust across different topological views, effectively aligning the pre-training and inference distributions.

By combining the interference-aware Transformer architecture with the hybrid self-supervised pre-training strategy, we construct a unified framework to effectively address the objective heterogeneity in wireless resource allocation and enhance few-shot generalization.
In summary, the main contributions of this paper are as follows.
\begin{itemize}
    \item \textbf{Graph foundation model for wireless resource allocation:} We propose GFM-RA, a graph foundation model for resource allocation that brings a self-supervised pre-training and downstream fine-tuning paradigm to wireless networks. By shifting from task-specific learning to general-purpose representation learning, GFM-RA significantly improves sample efficiency and enables highly efficient adaptation to heterogeneous downstream tasks with minimal additional fine-tuning. 

    \item \textbf{Interference-aware graph Transformer architecture:} We develop an interference-aware graph Transformer backbone for wireless resource allocation. To overcome the topological blindness of standard attention mechanisms, we introduce a novel bias projector that explicitly injects physical edge features into the attention scores. This design ensures that the global information aggregation is strictly grounded in the physics of wireless interference.

    \item \textbf{Hybrid self-supervised pre-training strategy:} We propose a robust pre-training paradigm that synergizes generative and contrastive learning without relying on expensive solver-generated labels. Specifically, we combine masked edge prediction to compel the model to reconstruct local interference patterns and a negative-free Teacher-Student contrastive learning mechanism to ensure global representation consistency against topological perturbations.

    \item \textbf{Empirical validation of scaling and generalization:} Extensive experiments demonstrate that GFM-RA achieves state-of-the-art performance. Crucially, we validate the scalability of our architecture compared to MPNNs, demonstrating that performance benefits from increased model depth without suffering from over-smoothing. Furthermore, the developed foundation model exhibits exceptional few-shot generalization capabilities in out-of-distribution (OOD) scenarios, verifying its potential as a general-purpose interference manager for future wireless networks.
\end{itemize}

The rest of the paper is organized as follows. Section II introduces the system model for wireless interference networks and formulates the resource allocation optimization problem under diverse utility objectives. Section III details the proposed foundation model, GFM-RA, elaborating on the interference-aware Transformer architecture and the hybrid self-supervised pre-training strategy. Section IV presents extensive simulation results to validate the superiority of the proposed method in terms of scalability, sample efficiency, and few-shot generalization. Finally, Section V concludes the paper.

\section{System Model and Problem Formulation}

We consider the problem of power control within a wireless interference network comprising $K$ mutually interfering transmitter-receiver pairs operating over a shared spectrum band. The extension to link scheduling for these transmitter-receiver pairs is straightforward.  We assume a block-fading channel model where channel states remain constant during one scheduling slot but vary independently between slots. Consequently, we focus on optimizing resource allocation for individual network snapshots based on the current channel realization.

Let $h_{kj} \in \mathbb{C}$ denote the complex channel gain from the $j$-th transmitter to the $k$-th receiver. Accordingly, $h_{kk}$ represents the direct link channel, while $h_{kj}$ ($j \neq k$) denotes the cross-link interference. The transmit power for link $k$, denoted by $p_k$, is constrained by a maximum budget $P_{\max}$, with the global configuration denoted by $\boldsymbol{p} = [p_1, \dots, p_K]^T$. The signal-to-interference-plus-noise ratio (SINR) at the $k$-th receiver is given by
\begin{equation}
\text{SINR}_k(\boldsymbol{p}) = \frac{|h_{kk}|^2 p_k}{\sum_{j \neq k} |h_{kj}|^2 p_j + \sigma^2},
\end{equation}
where $\sigma^2$ denotes the additive white Gaussian noise (AWGN) power. The achievable spectral efficiency (SE) for user $k$ is given by
\begin{equation}
R_k(\boldsymbol{p}) = \log_2(1 + \text{SINR}_k(\boldsymbol{p})).
\end{equation}

In practical wireless systems, user demands are highly dynamic and heterogeneous, causing the desired network performance metrics to frequently shift between sum-rate maximization and strict user fairness. To capture this objective heterogeneity within a unified mathematical framework, we formulate a flexible resource allocation problem. Specifically, the task is to maximize a system-level utility function subject to per-user power constraints. Consistent with the formulation in \cite{chowdhury2021unfolding}, this optimization task is defined as
\begin{equation}
\label{eq:generic_prob}
    \begin{aligned}
    \underset{\boldsymbol{p}}{\text{maximize}} \quad & \sum_{k=1}^{K} \beta(R_k(\boldsymbol{p})) \\
    \text{subject to} \quad & 0 \leq p_k \leq P_{\max}, \quad \forall k \in \{1, \dots, K\},
    \end{aligned}
\end{equation}
where $\beta(\cdot)$ represents a strictly increasing utility function. To address diverse network performance requirements, we consider three distinct utility forms:

\begin{itemize}
    \item \textbf{Sum Rate Maximization:} Defined by $\beta(z) = z$, this objective aims to maximize the aggregate network throughput. However, it tends to allocate resources disproportionately to users with strong channels, often causing starvation for cell-edge users.
    
    \item \textbf{PF Maximization:} Defined by $\beta(z) = \log(z)$, this metric maximizes the geometric mean of user rates. It provides a balanced trade-off, ensuring user fairness while maintaining a reasonable level of total sum rate.
    
    \item \textbf{QoS-Aware Optimization:} To enforce a minimum rate requirement $R_{\min}$ for each user, we employ a penalty-based formulation. The utility is designed as $\beta(z) = z - \alpha \cdot \max(0, R_{\min} - z)$, where the penalty factor $\alpha > 1$ penalizes QoS violations and is a design parameter in practice. This composite objective encourages high sum rates when QoS constraints are met but imposes a steep linear penalty for violations, effectively forcing the optimizer to prioritize the connectivity of disadvantaged users.
    
\end{itemize}

In general, problem \eqref{eq:generic_prob} is non-convex due to the coupled interference terms in the SINR expressions and is difficult to solve optimally in real time for large networks.
To exploit the relational structure of interference, we represent each instantaneous network realization as a fully connected directed graph $\mathcal{G} = (\mathcal{V}, \mathcal{E})$. Specifically, the node set $\mathcal{V} = \{v_1, \dots, v_K\}$ corresponds to the $K$ transmitter-receiver pairs, where each node $v_k$ is associated with the direct link gain $h_{kk}$. The edge set $\mathcal{E}$ captures the interference coupling between links. To align with the physical signal propagation model, we define a directed edge from node $v_j$ to node $v_k$ (for $j \neq k$) as $\mathbf{e}_{kj}$ weighted by the cross-link channel gain $h_{kj}$. Under this representation, the incoming edges to node $v_k$ aggregate the interference term $\sum_{j \neq k} |h_{kj}|^2 p_j$ in the SINR expression of user $k$. Therefore, the graph structure provides a natural abstraction of the interference topology that governs the utility in \eqref{eq:generic_prob}.

The goal of this work is to develop a foundation model for wireless resource allocation, pre-trained on large-scale datasets, which can adapt to various downstream tasks with minimal additional fine-tuning.  Conventional learning-based approaches typically train separate models for different objectives, which limits adaptability when the system-level utility changes. In contrast, our foundation model learns general-purpose features and representations of the underlying interference structure from unlabeled network instances, so that downstream adaptation to new utility functions can be achieved with limited task-specific data.

\section{Graph Foundation Model}

\begin{figure}[t]
    \centering
    \includegraphics[width=0.49\textwidth]{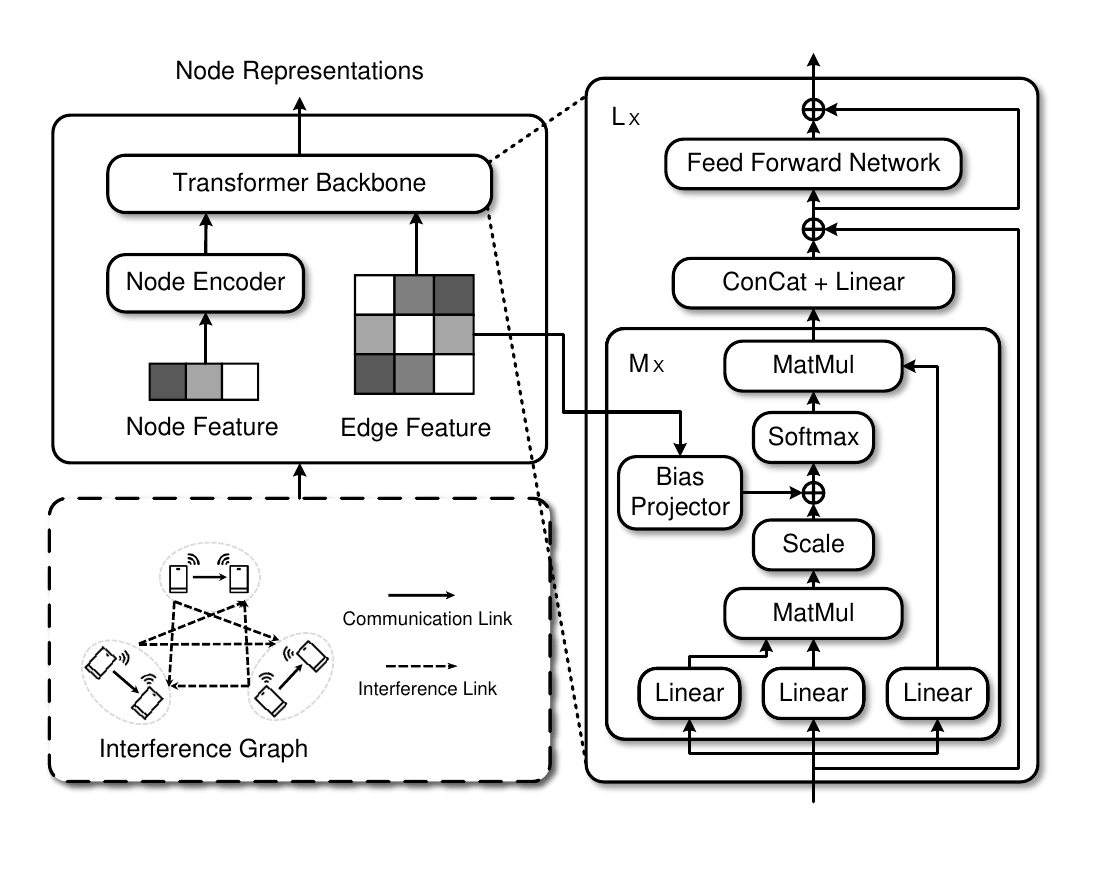}
    \caption{Architecture of the proposed foundation model, GFM-RA. The framework incorporates a bias projector within the graph foundation model block (right panel) to explicitly inject edge features (interference) into the attention mechanism for node representation learning.}
    \label{fig:Graphormer_encoder}
\end{figure}

To address the limitations of conventional optimization and task-specific learning models, we propose GFM-RA, a graph foundation model built on an interference-aware Transformer backbone and trained with a hybrid self-supervised pre-training strategy. As depicted in Fig.~\ref{fig:Graphormer_encoder}, the framework first encodes an instantaneous wireless network as an interference graph, then applies a graph Transformer to learn node representations that capture both local link states and global interference structure. These representations are pre-trained on unlabeled network instances and later adapted to downstream resource management tasks.

\subsection{Interference-Aware Graph Foundation Model}

Conventional MPNNs face a fundamental dilemma in wireless modeling: they rely on local aggregation which fails to capture long-range interference, while increasing network depth to expand the receptive field can induce over-smoothing, causing node embeddings to become indistinguishable \cite{li2018deeper, chen2020measuring}. This bottleneck limits effective scaling. To resolve this conflict, we transition from local message passing to a global interaction paradigm by designing a graph foundation model based on the Transformer architecture. Moreover, we introduce a bias projector that injects interference-graph edge features into attention scores, enabling topology-aware global aggregation while preserving physical interference structure.

\textbf{Feature Embedding.}
Recall that the node feature for $v_k$ represents the direct link state characterized by $h_{kk}$, while the edge feature $\mathbf{e}_{kj}$ is composed of the interference channel pair $[h_{kj}, h_{jk}]$, inspired by \cite{shen2022graph}. We first project the node feature into a high-dimensional latent space using a node encoder implemented as a two-layer MLP. This process yields the initial node representation $\mathbf{z}_k^{(0)} \in \mathbb{R}^{d_{\text{model}}}$.

\textbf{Bias-Injected Multi-Head Attention.}
The core innovation lies in the modification of the self-attention mechanism within the graph foundation model, as detailed in Fig.~\ref{fig:Graphormer_encoder}. To model heterogeneous interference patterns, we employ a multi-head attention architecture comprising $M$ parallel heads. Unlike standard Transformers that rely exclusively on the semantic similarity between node queries and keys, our approach explicitly incorporates physical interference information into the attention scores.

Specifically, a bias projector $\phi_B(\cdot)$ maps the edge features $\mathbf{e}_{kj}$ into a bias vector in $\mathbb{R}^M$, where the $m$-th element provides a head-specific bias. For head $m \in \{1, \dots, M\}$, the attention coefficient at layer $l$ is formulated as
\begin{equation}
    A_{kj}^{(l, m)} = \frac{(\mathbf{z}_k^{(l)}\mathbf{W}_{Q,m})(\mathbf{z}_j^{(l)}\mathbf{W}_{K,m})^T}{\sqrt{d_m}} + \phi_B^{(m)}(\mathbf{e}_{kj}),
\end{equation}
where $\mathbf{W}_{Q,m}, \mathbf{W}_{K,m} \in \mathbb{R}^{d_{\text{model}} \times d_m}$ are the learnable projection matrices for the $m$-th head, and $d_m = d_{\text{model}}/M$ is the dimension per head. By integrating the scalar bias $\phi_B^{(m)}(\mathbf{e}_{kj})$, we introduce a structural prior that compels some attention heads to prioritize dominant interference sources, while enabling the remaining heads to encode the broader network topology.

The outputs from all $M$ heads are then concatenated and projected to form the aggregated representation as
\begin{equation}
    \mathbf{z}_k^{(l)'} = \sum_{j \in \mathcal{V}} \left( \text{Concat}_{m=1}^{M} \left[ \text{Softmax}_j\left(A_{kj}^{(l, m)}\right) \left(\mathbf{z}_j^{(l)}\mathbf{W}_{V,m}\right) \right] \right) \mathbf{W}_O,
\end{equation}
where $\mathbf{W}_{V,m}$ and $\mathbf{W}_O$ denote the value and output projection matrices, respectively. This intermediate vector $\mathbf{z}_k^{(l)'}$ is subsequently processed by a feed-forward network with residual connections and layer normalization to produce the updated representation $\mathbf{z}_k^{(l+1)}$. Upon completing $L$ layers, the backbone yields the final node representations $\mathbf{Z}=[\mathbf{z}_1^{(L)}, \dots, \mathbf{z}_K^{(L)}]$, which serve as the generic embeddings for diverse downstream resource allocation tasks.

\begin{figure*}[t]
    \centering
    \includegraphics[width=1\textwidth]{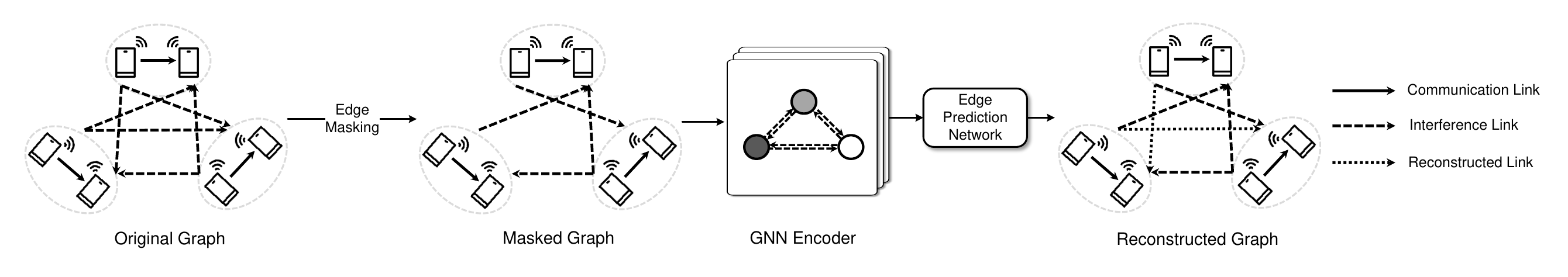}
    \caption{Illustration of the generative pre-training strategy based on edge prediction. The encoder reconstructs masked interference links from the partial graph topology to capture local interference patterns.}
    \label{fig:Edge_prediction}
\end{figure*}

\subsection{Hybrid Self-Supervised Pre-training}

Training a foundation model necessitates a self-supervised strategy to learn general-purpose features and representations of the network interference without relying on specific design objectives. To this end, we propose a hybrid optimization framework that synergizes generative and contrastive learning. In the following, we will first describe the details of the two learning paradigms, respectively, and then put forward the overall training procedure integrating these two paradigms. 

\textbf{Generative Learning: Masked Edge Prediction.}
As illustrated in Fig.~\ref{fig:Edge_prediction}, the generative training is formulated as a masked edge prediction task. The goal is to train the backbone to recover missing interference features from the surrounding graph context. Since interference relationships are encoded in edge attributes rather than only in node features, this task directly encourages the model to learn the structural dependencies among interfering links that govern wireless resource allocation. The masked edge prediction procedure consists of four steps: masking, tokenization, encoding, and reconstruction.

First, a partial view of the network topology is generated by randomly sampling a subset of interference links $\mathcal{E}_{\text{mask}} \subset \mathcal{E}$ according to a uniform masking ratio $\rho$. The remaining unmasked edges retain their original physical features to provide the necessary context for reconstruction.

Second, we apply a token replacement strategy where the original features $\mathbf{e}_{kj}$ of all masked links are substituted with a learnable vector $\boldsymbol{\epsilon}_{\text{mask}}$. This mechanism prevents the model from directly observing the ground-truth interference values while maintaining the underlying graph connectivity.

Third, the modified graph structure $\tilde{\mathcal{G}}$ is fed into the graph foundation model architecture described in the Section IV-B , collectively denoted as $f_\Phi(\cdot)$, parameterized by $\Phi$. The model encodes information over the mutually interfering links to generate the contextualized node representations, given by
\begin{equation}
    \mathbf{Z} = f_{\Phi}(\tilde{\mathcal{G}}),
\end{equation}
where $\mathbf{Z} = [\mathbf{z}_1^{(L)}, \dots, \mathbf{z}_K^{(L)}]$ denotes the aggregate set of embeddings produced after $L$ layers. By integrating the learnable mask bias $\boldsymbol{\epsilon}_{\text{mask}}$ into the attention mechanism, the resulting embeddings implicitly encode the structural voids within the network topology.

Finally, to recover the latent physical attributes, we introduce a lightweight edge decoder $d_{\Psi}(\cdot)$, parameterized by $\Psi$. This decoder takes in the representations of transmitter node $\mathbf{z}_j$ and the receiver node $\mathbf{z}_k$ to predict the missing interference feature. The predicted edge feature is computed as
\begin{equation}
    \hat{\mathbf{e}}_{kj} = d_{\Psi}(\mathbf{z}_k, \mathbf{z}_j).
\end{equation}
The pre-training objective is to minimize the reconstruction error over the masked set, defined as
\begin{equation}
    \mathcal{L}_{\text{edge}} = \sum_{(k,j) \in \mathcal{E}_{\text{mask}}} \| \mathbf{e}_{kj} - \hat{\mathbf{e}}_{kj} \|^2. \label{eq:edge_prediction}
\end{equation}
This objective encourages the encoder to learn representations that are informative enough to recover missing interference relationships from partial observations. To accurately reconstruct $\mathbf{e}_{kj}$, the model must perform high-level reasoning over multi-hop paths in the unmasked context, effectively capturing the underlying interference patterns embedded in the wireless network topologies.


\begin{figure*}[t]
    \centering
    \includegraphics[width=1\textwidth]{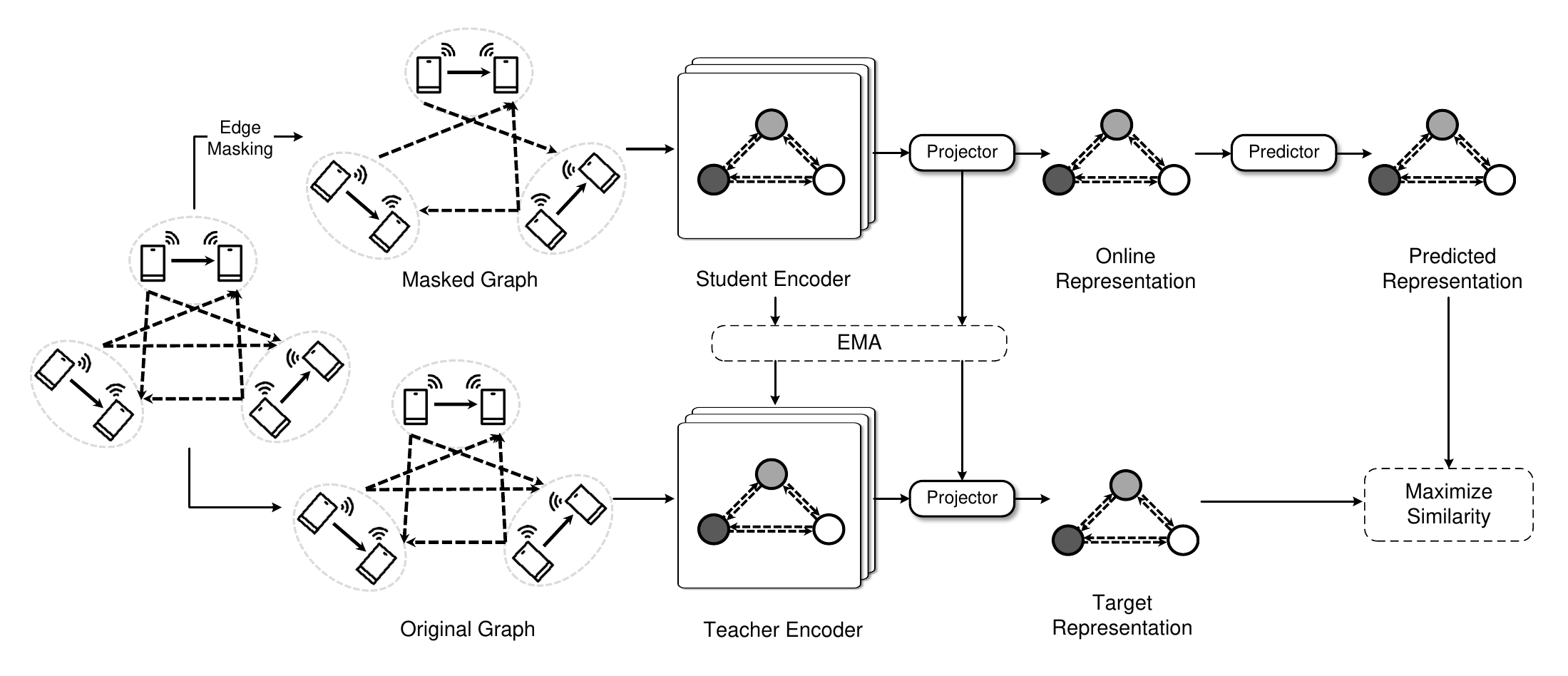}
    \caption{The proposed contrastive pre-training framework utilizing a Teacher-Student architecture. The Student encoder predicts the target representations of the Teacher encoder, updated via EMA, to ensure robustness against topological variations.}
    \label{fig:Contrastive_learning}
\end{figure*}

\textbf{Contrastive Learning: Teacher-Student Consistency.}
While the generative edge prediction training compels the model to capture network interference structures, relying on reconstruction alone may not be sufficient for downstream wireless resource allocation tasks. It is also important to impose representation consistency between the masked graphs and the original complete graphs such that the model can learn a general-purpose and robust representation of the underlying interference networks. As a result, we introduce a negative-free contrastive learning mechanism that promotes representation consistency across different views of the same interference network.

As depicted in Fig.~\ref{fig:Contrastive_learning}, the framework utilizes an asymmetric Teacher-Student architecture. The Teacher branch processes the unmasked, complete graph to provide a stable target representation of the complete interference network. Meanwhile, the Student branch operates on the masked graph and is optimized to align its predicted embeddings with the target representations generated by the Teacher. By forcing the Student to align with this reliable target, the model learns to better capture the global interference topology driven by dominant links. Consequently, the generated node embeddings become robust and insensitive to the random missing edges caused by masking.

Specifically, the Student network processes the online view, defined as the masked graph $\tilde{\mathcal{G}}$, using the foundation model backbone described in Section III-A, which is denoted as $f_{\theta_s}(\cdot)$ with parameters $\theta_s$ and follows the same steps as introduced in generative learning. This yields the Student node embeddings $\mathbf{Z}_s = f_{\theta_s}(\tilde{\mathcal{G}})$. In contrast, the Teacher network processes the target view, defined as the original unmasked graph $\mathcal{G}$, using the same foundation model backbone, but with a different set of parameters $\theta_t$. This yields the Teacher node embeddings $\mathbf{Z}_t = f_{\theta_t}(\mathcal{G})$.

Next, we map these backbone features into a metric space to align their representations. In the Student branch, the embeddings $\mathbf{Z}_s$ are mapped by a projector network $g_{\xi}(\cdot)$, parameterized by $\xi$, to produce the online representation, denoted as $\mathbf{v}_s = g_{\xi}(\mathbf{Z}_s)$. To prevent the collapse of representations into trivial constant solutions (e.g., where both the Student and Teacher trivially output identical all-zero vectors to minimize distance), we introduce an additional predictor network $q_{\omega}(\cdot)$, parameterized by $\omega$. This predictor transforms the online representation into the final predicted representation $\mathbf{u}_s$. The complete forward path for the Student branch is formulated as
\begin{equation}
    \mathbf{u}_s = q_{\omega}(g_{\xi}(f_{\theta_s}(\tilde{\mathcal{G}}))).
\end{equation}
Simultaneously, in the Teacher branch, the embeddings $\mathbf{Z}_t$ are also mapped by a projector network, which shares the same architecture as the Student branch, but with a different set of parameters $\xi'$, to produce the target representation $y_t$, expressed as
\begin{equation}
    \mathbf{y}_t = g_{\xi'}(f_{\theta_t}(\mathcal{G})).
\end{equation}
Note that the Teacher branch does not employ the predictor network that is symmetric with the Student branch. 

Consistency is enforced through parameter evolution and similarity maximization. Unlike the Student parameters $(\theta_s, \xi, \omega)$ which are updated via backpropagation, the Teacher parameters $(\theta_t, \xi')$ are not trained directly. Instead, they serve as a stable regression target and evolve through an exponential moving average (EMA) of the Student parameters to ensure training stability, expressed as
\begin{equation}\label{eq:EMA}
    \theta_t \leftarrow \tau\theta_t + (1-\tau)\theta_s, \quad \xi' \leftarrow \tau\xi' + (1-\tau)\xi,
\end{equation}
)where $\tau \in (0, 1)$ is the momentum coefficient. The training objective is to maximize the cosine similarity between the Student's predicted representation $\mathbf{u}_s$ and the Teacher's target representation $\mathbf{y}_t$, defined as
\begin{equation}
    \mathcal{L}_{\text{cl}} = - \frac{1}{|\mathcal{V}|} \sum_{k \in \mathcal{V}} \frac{\mathbf{u}_s(k)^T \mathbf{y}_t(k)}{\|\mathbf{u}_s(k)\| \|\mathbf{y}_t(k)\|}. \label{eq:contrastive_loss}
\end{equation}
By minimizing this loss, the backbone $f_{\theta_s}(\cdot)$  learns to distill the essence of the complete topology $\mathcal{G}$ albeit observing only the partial evidence $\tilde{\mathcal{G}}$.

\textbf{Hybrid Pre-training.}
To exploit the advantages of both local link reconstruction and global representation invariance, we integrate the generative and contrastive paradigms into a unified optimization framework. This hybrid strategy is designed to mitigate the limitations inherent in using either approach in isolation. The masked edge prediction term encourages the model to recover missing interference relationships from partial observations, while the Teacher-Student consistency term regularizes the learned representations to remain stable across different views of the same graph. The total pre-training objective is constructed as a weighted linear combination of the masked edge prediction loss and the Teacher-Student consistency loss. Letting $\Theta$ denote the complete set of trainable parameters within the backbone and the projection heads, the final optimization target is defined as
\begin{equation}
    \mathcal{L}_{\text{pre}}(\Theta) = \mathbb{E}_{\mathcal{G}} \left[  \mathcal{L}_{\text{edge}} + \lambda \mathcal{L}_{\text{cl}}\right],
\end{equation}
where the hyperparameter $\lambda$ acts as a balancing coefficient that modulates the contribution of the two components.  In practice, this objective is optimized using stochastic gradient descent over mini-batches of graphs and randomly sampled masking patterns. The complete hybrid pre-training procedure is summarized in Algorithm~\ref{alg:pretraining}.

\begin{algorithm}[t]
\caption{Hybrid Self-Supervised Pre-training Strategy}
\label{alg:pretraining}
\SetAlgoLined
\DontPrintSemicolon

\SetKwBlock{MyBlock}{}{}
\newcommand{\StepBlock}[2]{
    \textbf{#1} 
    \MyBlock{   
        #2
    }
}

\KwIn{Dataset $\mathcal{D}$, Masking ratio $\rho$, EMA rate $\tau$, Balance weight $\lambda$, Learning rate $\eta$.}
\KwOut{Pre-trained Student parameters $\Theta_s$.}

Initialize Student parameters $\Theta_s= \{\theta_s, \xi, \omega, \Psi\}$ randomly.\;
Initialize Teacher parameters $\Theta_t= \{\theta_t, \xi'\}$ as a copy of Student.\;

\While{not converged}{
    Sample a mini-batch of graphs $\mathcal{B} \subset \mathcal{D}$.\;
    
    \For{each graph $\mathcal{G} \in \mathcal{B}$}{
        \StepBlock{1) View Construction:}{
            Sample mask indices $\mathcal{E}_{\text{mask}} \subset \mathcal{E}$ with ratio $\rho$.\;
            Generate online view $\tilde{\mathcal{G}}$ by replacing $\mathbf{e}_{kj}$ with $\boldsymbol{\epsilon}_{\text{mask}}$ for all $(k,j) \in \mathcal{E}_{\text{mask}}$.\; 
            Let target view be $\mathcal{G}$.\;
        }
        
        \StepBlock{2) Forward Propagation:}{
            Obtain Student node embeddings: $\mathbf{Z}_s = f_{\theta_s}(\tilde{\mathcal{G}})$.\; 
            Obtain Teacher node embeddings: $\mathbf{Z}_t = f_{\theta_t}(\mathcal{G})$.\;
        }
        
        \StepBlock{3) Generative Branch:}{
            Predict masked edges: $\hat{\mathbf{e}}_{kj} = d_{\Psi}(\mathbf{z}_{s,k}, \mathbf{z}_{s,j}), \ \forall (k,j) \in \mathcal{E}_{\text{mask}}$.\;
            Compute reconstruction loss $\mathcal{L}_{\text{edge}}$ via \eqref{eq:edge_prediction}.\;
        }
        
        \StepBlock{4) Contrastive Branch:}{
            Student predicted representation: $\mathbf{u}_s = q_{\omega}(g_{\xi}(\mathbf{Z}_s))$.\;
            Teacher target representation: $\mathbf{y}_t = g_{\xi'}(\mathbf{Z}_t)$.\;
            Compute consistency loss $\mathcal{L}_{\text{cl}}$ via \eqref{eq:contrastive_loss}.\;
        }
    }
 
    Calculate total pre-training loss:\;
    \Indp \Indp \Indp
    $\displaystyle \mathcal{L}_{\text{pre}} = \frac{1}{|\mathcal{B}|} \sum_{\mathcal{G} \in \mathcal{B}} (\mathcal{L}_{\text{edge}} + \lambda \mathcal{L}_{\text{cl}}).$\;
    \Indm \Indm \Indm
    
    Update Student: $\Theta_s \leftarrow \Theta_s - \eta \nabla \mathcal{L}_{\text{pre}}$.\;
    Update Teacher via \eqref{eq:EMA}.\;
}
\end{algorithm}

\subsection{Fine-tuning Foundation Models to Downstream Tasks}

In this section, we leverage the pre-trained foundation model to address a range of wireless downstream tasks, such as allocating wireless resources to optimize vastly different utilities defined in Section II. For each task, we attach a lightweight decision head, e.g., a two-layer MLP, to the task-agnostic foundation model and then perform fine-tuning on a limited dataset. 
In this case, the decision head maps the final node representation $\mathbf{z}_k^{(L)}$ to a normalized power action. To strictly enforce the power constraint $0 \leq p_k \leq P_{\max}$, we employ a Sigmoid activation scaled by $P_{\max}$, expressed as
\begin{equation}
    p_k = P_{\max} \cdot \text{Sigmoid}(\text{MLP}(\mathbf{z}_k^{(L)})).
\end{equation}

The fine-tuning loss function for each task is the negative of the expected sum utility, formulated as
\begin{equation}
    \mathcal{L}_{\text{down}}(\theta) = - \mathbb{E}_{\mathcal{G}} \left[ \sum_{k=1}^{K} \beta\left(R_k(\boldsymbol{p}) \right) \right].
\end{equation}

To effectively adapt the general-purpose representations to this specific objective without destroying the pre-trained structural knowledge, we implement a two-stage optimization strategy.
In the first stage, referred to as \textit{Head Warmup}, we freeze the parameters of the graph foundation model backbone and exclusively update the decision head. This precautionary step prevents the backpropagation of large, noisy gradients from the untrained head which could otherwise destabilize the well-learned topological features.
In the second stage, referred to as \textit{Full Fine-tuning}, we unfreeze the backbone parameters and jointly optimize the entire network using a substantially reduced learning rate. This phase allows the graph foundation model structure to perform fine-grained adjustments to the node embeddings, thereby effectively aligning the semantic space with the nuances of the downstream utility maximization task.

\section{Simulation Results}

\subsection{Simulation Settings}
\subsubsection{Dataset Generation}
To comprehensively assess the robustness and transferability of the proposed framework, we established a high-fidelity simulation environment for device-to-device underlay networks within a 1,000 $\times$ 1,000$~m^2$ square region. The experimental evaluation relies on a curated library of 20 distinct datasets, indexed as $\mathcal{D}_1$ to $\mathcal{D}_{20}$, which are categorized into pre-training and few-shot adaptation tasks.

\textbf{Simulation Environment.} 
The network topology is constructed by uniformly distributing paired receivers within an annular region $[d_{\min}, d_{\max}]$ centered at their respective transmitters. To ensure valid user association, a strict nearest-neighbor constraint is enforced: a receiver location is retained solely if it is geographically closer to its serving transmitter than to any interfering node; otherwise, the position is regenerated. The wireless channel is modeled as a superposition of small-scale Rayleigh fading and large-scale dual-slope path loss with log-normal shadowing (standard deviation $\sigma_{\mathrm{sh}} = 7$ dB)~\cite{naderializadeh2023learning, zhang2015downlink}. The system operates over a bandwidth of $W = 10$ MHz, with a maximum transmit power budget of $P_{\max} = 10$ dBm and a noise power spectral density (PSD) of $-174$ dBm/Hz.

\textbf{Pre-training Datasets $\mathcal{D}_1$-$\mathcal{D}_{15}$.} 
To foster robust generalization across a continuum of interference regimes, we generate a heterogeneous collection of 15 datasets used exclusively for pre-training. These scenarios ($\mathcal{D}_1$ through $\mathcal{D}_{15}$) are formed by the cross-combination of five user density levels $K \in \{20, 35, 50, 65, 80\}$ and three varying transmitter-receiver distance ranges $[d_{\min}, d_{\max}] \in \{ [2, 65], [10, 50], [30, 70] \}$~m. This configuration covers environments ranging from noise-limited sparse networks to interference-limited ultra-dense clusters. 
Each dataset in this group contains 120,000 network snapshots, partitioned into 100,000 for training, 10,000 for validation, and 10,000 for testing.

\textbf{Fine-tuning and Evaluation Datasets $\mathcal{D}_{16}$-$\mathcal{D}_{20}$.} 
To evaluate the foundation model's capability to adapt to unseen distributions with limited data availability, we construct 5 additional datasets. These scenarios are characterized by a highly variable link distance range of $[d_{\min}, d_{\max}] = [1, 100]$~m, which poses a greater challenge than the pre-training distributions. The datasets correspond to the five user density levels $K \in$ \{20, 35, 50, 65, 80\}, respectively assigned to $\mathcal{D}_{16}$ through $\mathcal{D}_{20}$.
Unlike the pre-training phase, these datasets are designed to test sample efficiency. For each dataset $\mathcal{D}_{16}$-$\mathcal{D}_{20}$, we evaluate performance using a variable number of $N_{\text{shot}} \in $ \{64, 128, 256, 512, 1,024, 2,048\}, while the testing set size is held constant at 5,000 snapshots to ensure statistical reliability.

\subsubsection{Network Hyperparameters and Pre-training Details} 
As shown in Table~\ref{tab:hyperparameters}, the proposed GFM-RA backbone is instantiated with $L=6$ Transformer layers. The hidden embedding dimension is set to $d_{\text{model}} = 768$, and the multi-head attention mechanism employs $32$ parallel heads to capture diverse interference features. The decision head used for fine-tuning is a two-layer MLP with ReLU activation.

\textbf{Implementation Setup.}
All models are implemented using the PyTorch framework and PyTorch Geometric library. The hybrid pre-training phase is conducted for 200 epochs using the Adam optimizer, configured with a batch size of 512 and an initial learning rate of $1 \times 10^{-4}$. Regarding task-specific hyperparameters, we adopt a uniform masking ratio of $\rho = 0.3$ for the generative task, while the contrastive task utilizes an EMA decay rate of $\tau = 0.996$ and a loss balancing coefficient of $\lambda = 0.1$. To enhance convergence stability, we employ a dynamic scheduler that decays the learning rate by a factor of 0.5 following a validation stagnation patience of 10 epochs.

\textbf{Fine-tuning Details.}
The downstream adaptation follows the two-stage protocol consisting of a head warmup (10 epochs) and full fine-tuning (100 epochs). During full fine-tuning, we apply differential learning rates to balance stability and plasticity: the pre-trained backbone is optimized with a lower learning rate of $1 \times 10^{-4}$ to preserve topological knowledge, while the randomly initialized decision head employs a higher learning rate of $1 \times 10^{-3}$ for rapid convergence. Furthermore, specifically for the QoS-aware optimization task, the minimum rate requirement and the penalty factor are empirically set to $R_{\min} = 0.3$ bps/Hz and $\alpha = 15$, respectively, to strictly enforce fairness constraints during the adaptation.

\begin{table}[h]
    \centering
    \caption{List of Hyperparameters and Training Settings}
    \label{tab:hyperparameters}
    \renewcommand{\arraystretch}{1.2}
    \begin{tabular}{l|c|c}
        \hline
        \textbf{Category} & \textbf{Parameter / Description} & \textbf{Value} \\
        \hline
        \hline
        \multirow{3}{*}{Architecture} 
        & Number of Layers ($L$) & 6 \\
        & Hidden Dimension ($d_{\text{model}}$) & 768 \\
        & Number of Attention Heads ($M$) & 32 \\
        \hline
        \multirow{7}{*}{Pre-training} 
        & Total Epochs & 200 \\
        & Batch Size & 512 \\
        & Learning Rate ($\eta$) & $1 \times 10^{-4}$ \\
        & Masking Ratio ($\rho$) & 0.3 \\
        & EMA Decay Rate ($\tau$) & 0.996 \\
        & Loss Balance Weight ($\lambda$) & 0.1 \\
        & LR Scheduler Factor & 0.5 \\
        & LR Scheduler Patience & 10 epochs \\
        \hline
        \multirow{4}{*}{Fine-tuning} 
        & Head Warmup Epochs & 10 \\
        & Full Fine-tuning Epochs & 100 \\
        & Backbone Learning Rate & $1 \times 10^{-4}$ \\
        & Head Learning Rate & $1 \times 10^{-3}$ \\
        \hline
    \end{tabular}
\end{table}

\begin{figure*}[t]
    \centering
    \includegraphics[width=1\linewidth]{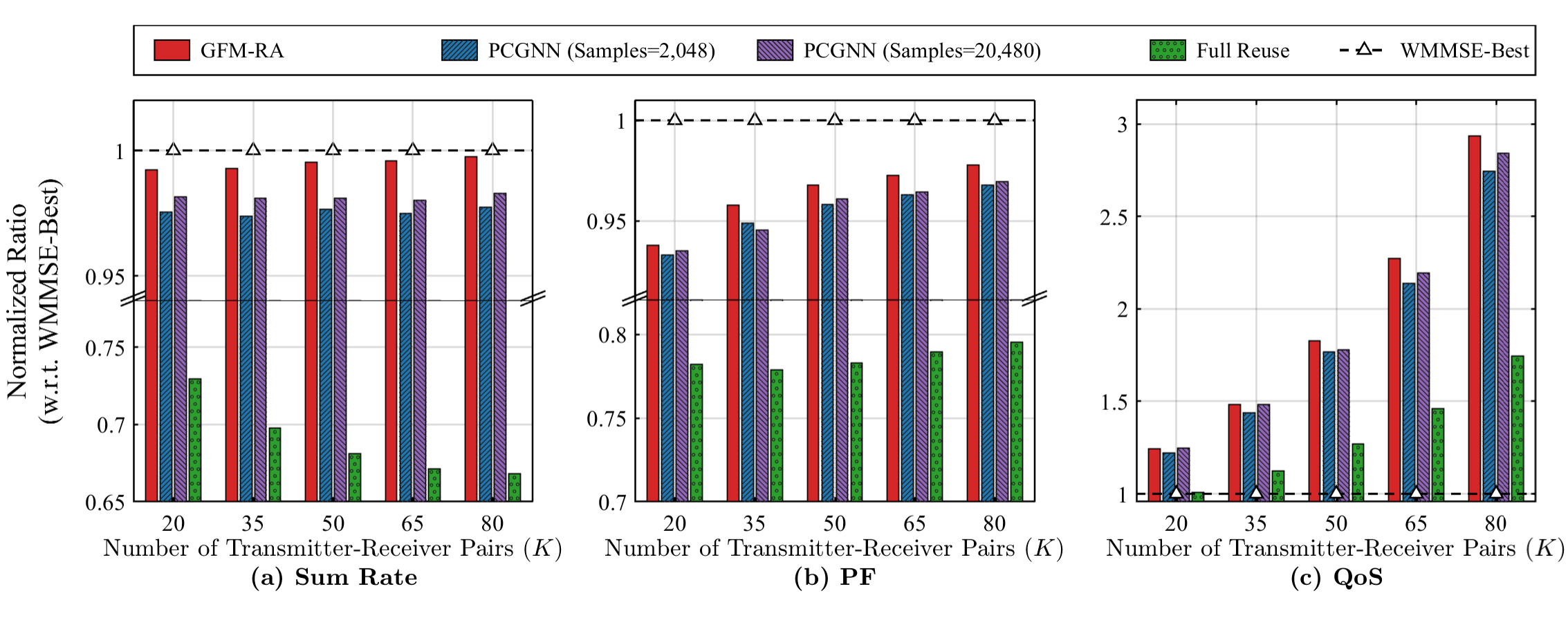}
    \caption{Performance of adapting the foundation model to three distinct utility functions under varying network densities in OOD scenarios ($\mathcal{D}_{16}$--$\mathcal{D}_{20}$).}
    \label{fig:multi_task}
\end{figure*}

\begin{figure*}[t]
    \centering
    \includegraphics[width=1\linewidth]{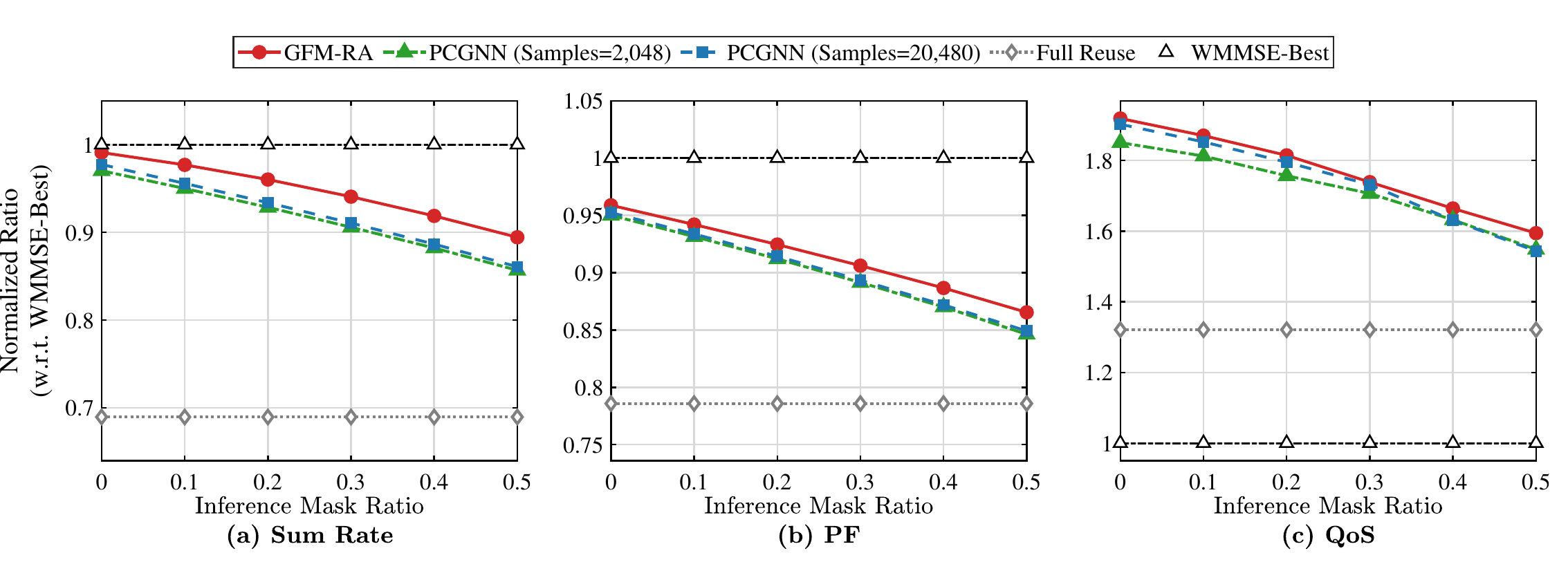}
    \caption{Robustness of the foundation model against the incompleteness of CSI. The aggregated performance is averaged across five OOD scenarios ($\mathcal{D}_{16}$--$\mathcal{D}_{20}$).}
    \label{fig:masked_inference}
\end{figure*}

\subsubsection{Baselines}

To rigorously assess the performance of the proposed framework, we benchmark it against a diverse set of baselines, ranging from heuristic schemes to advanced iterative optimization and DL approaches:
\begin{itemize}
    \item \textbf{WMMSE-Best \cite{christensen2008wmmse}:} As a representative of traditional iterative optimization, we employ the WMMSE algorithm~\cite{christensen2008wmmse}. Acknowledging the non-convex nature of the underlying problem, which renders the algorithm susceptible to local optima, we implement a robust multi-start strategy. Specifically, the algorithm is executed over 100 independent trials with random initializations, and the solution yielding the maximum utility is reported as the final result.
    
    \item \textbf{PCGNN \cite{shen2022graph}:} We utilize this model as the representative state-of-the-art for learning-based solutions. It relies on the standard MPNN paradigm to aggregate local neighborhood information. 
    
    \item \textbf{Full Reuse:} This static baseline assumes a non-cooperative environment where every transmitter operates at the maximum power budget $P_{\max}$.
\end{itemize}

\subsection{Performance Evaluation}

Fig.~\ref{fig:multi_task} presents a comprehensive performance evaluation across three downstream tasks, i.e., distinct utility functions: (a) Sum Rate, (b) PF, and (c) QoS. The experiments are conducted on the OOD datasets ($\mathcal{D}_{16}$-$\mathcal{D}_{20}$) across varying network densities, ranging from $K=20$ to $K=80$ transmitter-receiver pairs. The performance metric is defined as the normalized utility ratio relative to the WMMSE-Best benchmark. In this evaluation, the proposed foundation model GFM-RA is pre-trained in the datasets $\mathcal{D}_1$-$\mathcal{D}_{15}$ and fine-tuned using datasets $\mathcal{D}_{16}$-$\mathcal{D}_{20}$ with 2,048 samples in each scenario. To establish a rigorous comparison, the PCGNN baseline is evaluated under two distinct training regimes with datasets $\mathcal{D}_{16}$-$\mathcal{D}_{20}$: fine-tuned with 2,048 samples, denoted as ``PCGNN (Samples=2,048)'', and fine-tuned with an augmented dataset of 20,480 samples, denoted as ``PCGNN (Samples=20,480)''.

As illustrated in Fig.~\ref{fig:multi_task}, GFM-RA consistently outperforms the PCGNN (Samples=2,048) baseline across all optimization objectives and network scales. Notably, despite benefiting from a 10-fold increase in training data, the PCGNN (Samples=20,480) exhibits merely a marginal performance gain over its 2,048-sample counterpart. This saturation phenomenon indicates that the MPNN architecture is inherently bottlenecked, failing to capture higher-order topological dependencies in dense interference graphs and thus remaining strictly sub-optimal compared to our algorithm. 
Furthermore, as shown in Fig.~\ref{fig:multi_task}(a), GFM-RA exhibits highly competitive performance in the unconstrained sum rate maximization task, closely matching the high-quality solutions produced by the WMMSE-Best baseline. 
Furthermore, in the constraint-driven QoS task shown in Fig.~\ref{fig:multi_task}(c), the traditional WMMSE algorithm proves suboptimal in balancing aggregate throughput with user fairness. 
Conversely, GFM-RA vastly surpasses the traditional algorithms, achieving a normalized ratio nearly up to 3.0 at $K=80$. This demonstrates that the generalized embeddings extracted by GFM-RA can be seamlessly and efficiently re-purposed to navigate complex constraint landscapes, achieving superior multi-objective optimization that traditional iterative algorithms struggle to resolve.

Fig.~\ref{fig:masked_inference} examines the foundation model's robustness against the incompleteness of CSI, which usually relies on user feedback in practice. 
Specifically, we randomly drop some edges in the interference graph before feeding it to the model, or, equivalently, CSI corresponding to these edges is missing. We vary the ratio of links without CSI reports from 0 to 0.5 across five OOD scenarios ($\mathcal{D}_{16}$-$\mathcal{D}_{20}$). The plotted average performance is normalized against the WMMSE-Best baseline with perfect full CSI (unmasked graphs). 
As illustrated in Fig.~\ref{fig:masked_inference}, while performance inevitably degrades for all learning-based methods as information loss intensifies, GFM-RA consistently outperforms the baselines across all three optimization tasks. 
Notably, as shown in Fig.~\ref{fig:masked_inference}(a)-(b), increasing the training data for PCGNN tenfold, i.e., from 2,048 to 20,480 samples, yields negligible improvement in robustness. 
In contrast, even under the extreme condition where 50\% of the interference graph is unobservable,  GFM-RA maintains a high sum rate ratio of approximately 0.90, preserves a distinct superiority in PF, and sustains a staggering performance of over 1.6$\times$ the WMMSE-Best baseline in the QoS task. 
This structural robustness is inherently derived from the generative pre-training phase, which explicitly conditions the backbone to infer missing links from the available context, enabling the model to reconstruct the latent interference environment and make near-optimal decisions despite partial observability.

\subsection{Few-shot Performance Evaluation}
In this subsection, few-shot learning experiments are evaluated across the OOD datasets ($\mathcal{D}_{16}$--$\mathcal{D}_{20}$). For statistical reliability, the reported results represent the average performance across these distinct datasets. The comparative analysis involves two schemes: the proposed model trained from scratch, denoted as ``GFM-RA (From Scratch)'', and the proposed model initialized with pre-trained weights, denoted as ``GFM-RA (Pre-trained)''. The latter inherits parameters pre-trained with datasets $\mathcal{D}_1$--$\mathcal{D}_{15}$, thereby allowing for a direct evaluation of the pre-training strategy's efficacy. This experimental design specifically aims to quantify the foundation model's ability to facilitate rapid adaptation to unseen distributions under data-scarce conditions.

\begin{figure}[t]
    \centering
    \includegraphics[width=1\linewidth]{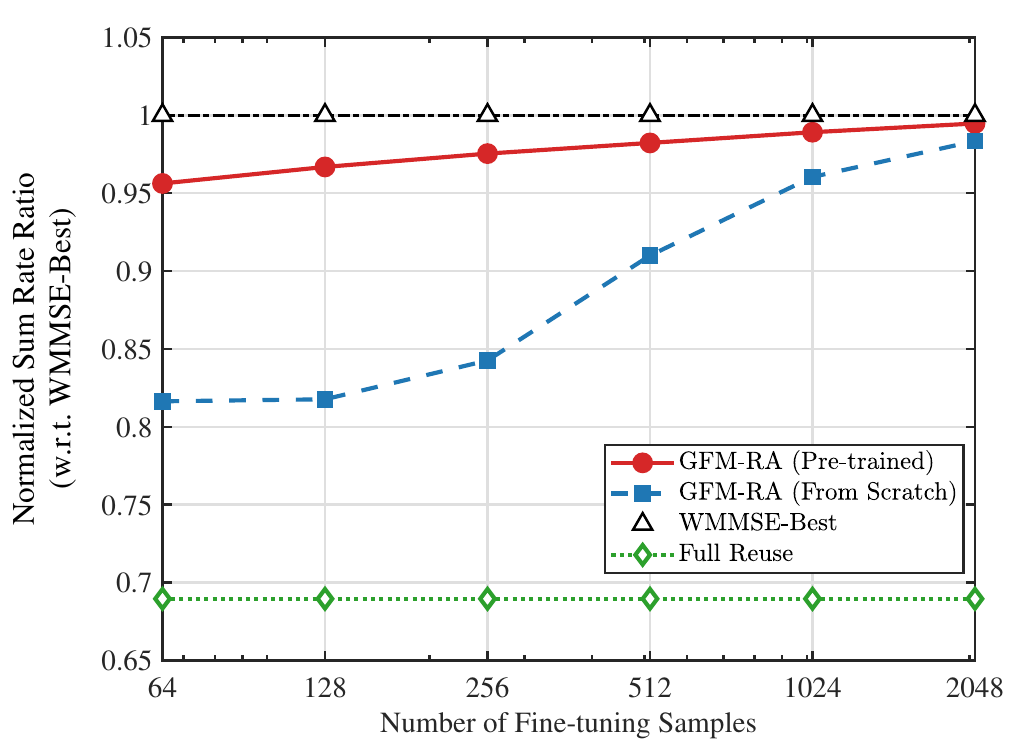}
    \caption{Few-shot adaptation performance for sum rate maximization across OOD scenarios.}
    \label{fig:fewshot_sumrate}
\end{figure}

Fig.~\ref{fig:fewshot_sumrate} evaluates the few-shot adaptation performance across OOD datasets ($\mathcal{D}_{16}$-$\mathcal{D}_{20}$). Results are reported as the normalized sum rate ratio relative to the WMMSE-Best benchmark, plotted against the number of fine-tuning samples.
The model trained from scratch, ``GFM-RA (From Scratch)'', exhibits a distinct ``cold-start'' phenomenon, with an initial normalized performance of 0.82 at 64 samples. This behavior corroborates the insight that Transformer-based architectures, lacking explicit graph inductive bias, require substantially larger datasets to infer topological dependencies. In contrast, the proposed pre-training strategy effectively circumvents this data-efficiency bottleneck. ``GFM-RA (Pre-trained)'' demonstrates superior adaptability, achieving a normalized sum rate exceeding 0.95 with merely 64 samples. The significant performance gap between the pre-trained and clean models indicates that the hybrid pre-training objective successfully embeds a generalized understanding of physical interference structures into the backbone. This physics-aware initialization facilitates robust knowledge transfer to unseen interference distributions, ensuring rapid convergence and near-optimal performance even in data-scarce regimes.

\begin{figure}[t]
    \centering
    \includegraphics[width=1\linewidth]{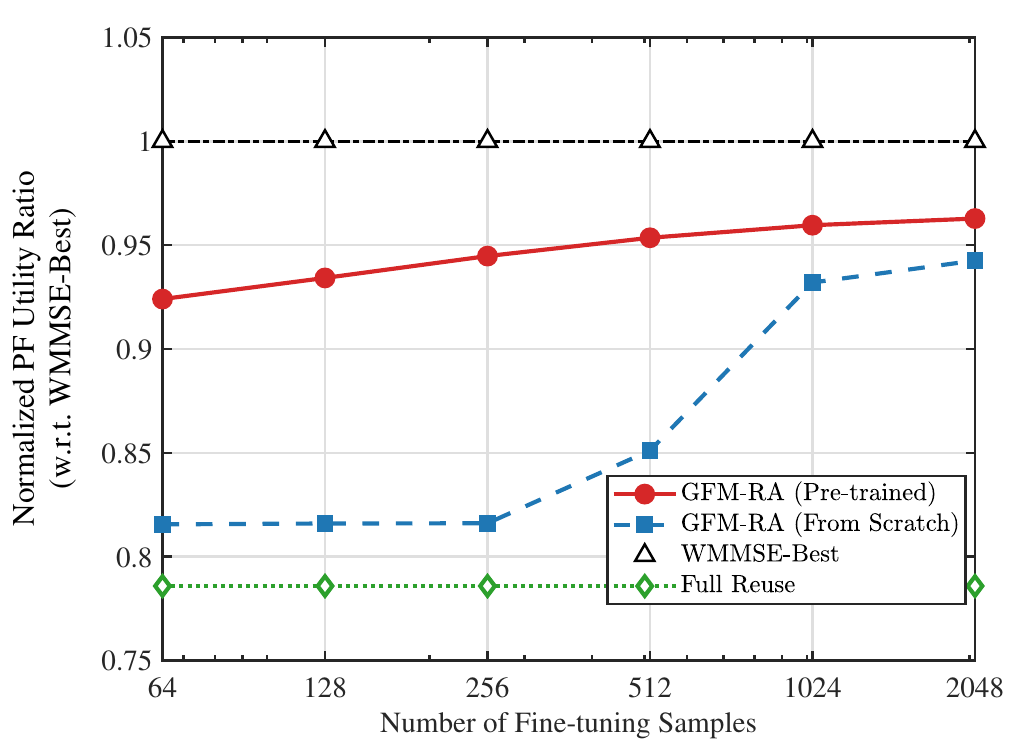}
    \caption{Few-shot adaptation performance for PF utility maximization in OOD scenarios.}
    \label{fig:fewshot_pf}
\end{figure}

Fig.~\ref{fig:fewshot_pf} extends the foundation model's adaptation analysis to the PF utility maximization task. Consistent with the sum rate analysis in Fig.~\ref{fig:fewshot_sumrate}, the pre-trained model successfully bypasses the severe cold-start limitations of the clean model, achieving over 92\% of the WMMSE-Best benchmark with merely 64 samples. This empirically validates the foundational hypothesis that the proposed self-supervised pre-training objectives, comprising masked edge prediction and contrastive consistency, capture the intrinsic topology of interference graphs instead of overfitting to specific downstream metrics.

\begin{figure*}[t]
    \centering
    \includegraphics[width=1\linewidth]{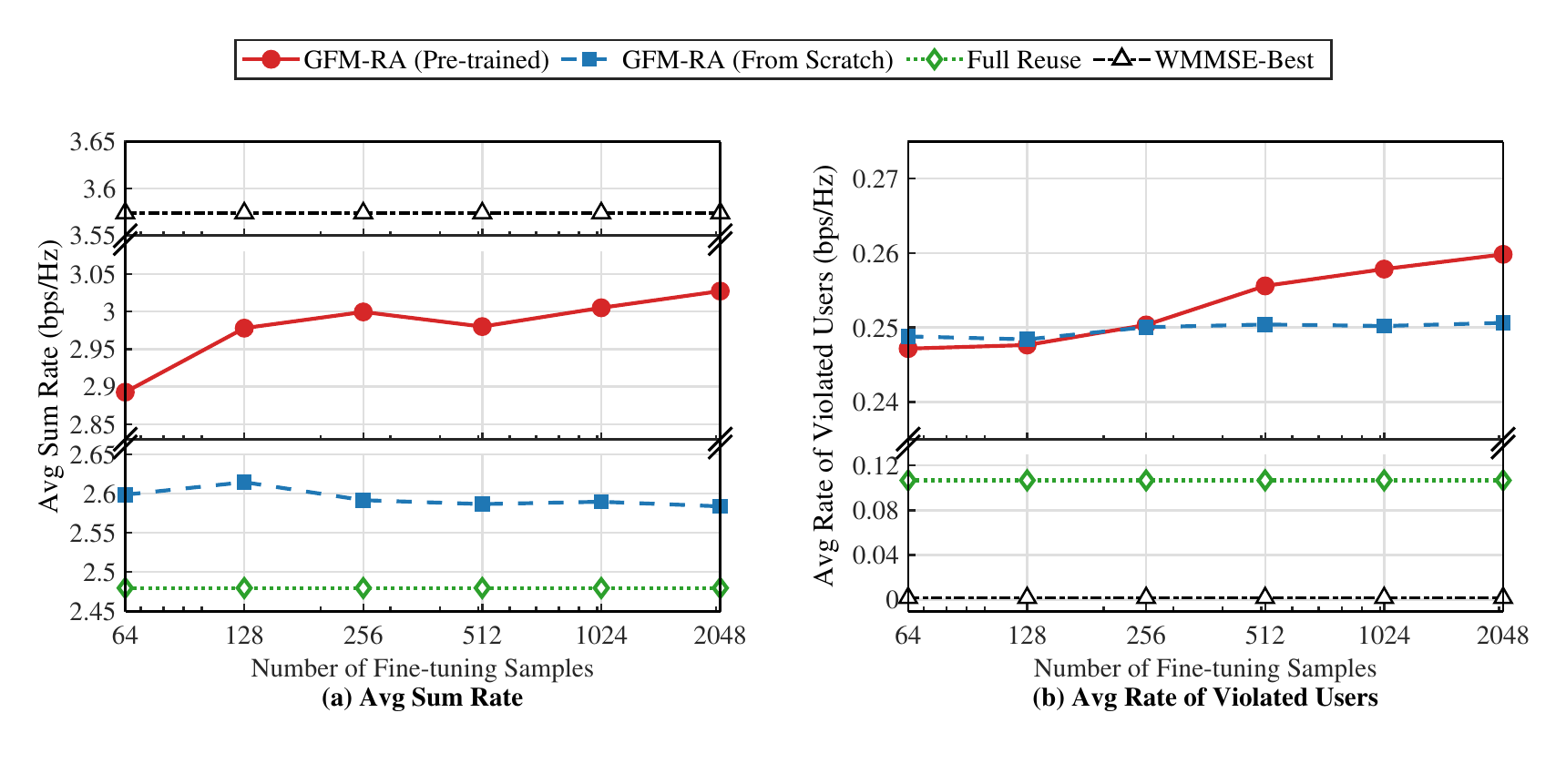}
    \caption{Few-shot adaptation performance for the QoS-aware objective in OOD scenarios: (a) average sum rate and (b) average rate of users violating the minimum threshold ($R_k < R_{\min}$).}
    \label{fig:qos_analysis}
\end{figure*}


Fig.~\ref{fig:qos_analysis} evaluates the QoS-aware performance in OOD scenarios using two metrics: (a) Average Sum Rate and (b) Average Rate of Violated Users, i.e., with rates $R_k < R_{\min}$. Formulated for unconstrained maximization, WMMSE establishes a theoretical upper bound for aggregate capacity but severely sacrifices vulnerable cell-edge users. In contrast, although ``GFM-RA (Pre-trained)'' achieves roughly 85\% of the WMMSE total sum rate, it drastically boosts the data rates of violated users, demonstrating a superior capability to protect vulnerable links.
Furthermore, the performance of ``GFM-RA (From Scratch)'' improves only marginally even as fine-tuning samples increase, while ``GFM-RA (Pre-trained)'' demonstrates rapid sample efficiency. With increasing data, ``GFM-RA (Pre-trained)'' swiftly learns to navigate the constrained solution space, accepting a necessary reduction in total aggregate rates to dramatically enhance the protection of violated users. This empirically validates that pre-trained physical knowledge ensures robust and rapid adaptation for complex constrained optimization.

\subsection{Scalability Analysis}

\begin{figure}[t]
    \centering
    \includegraphics[width=1\linewidth]{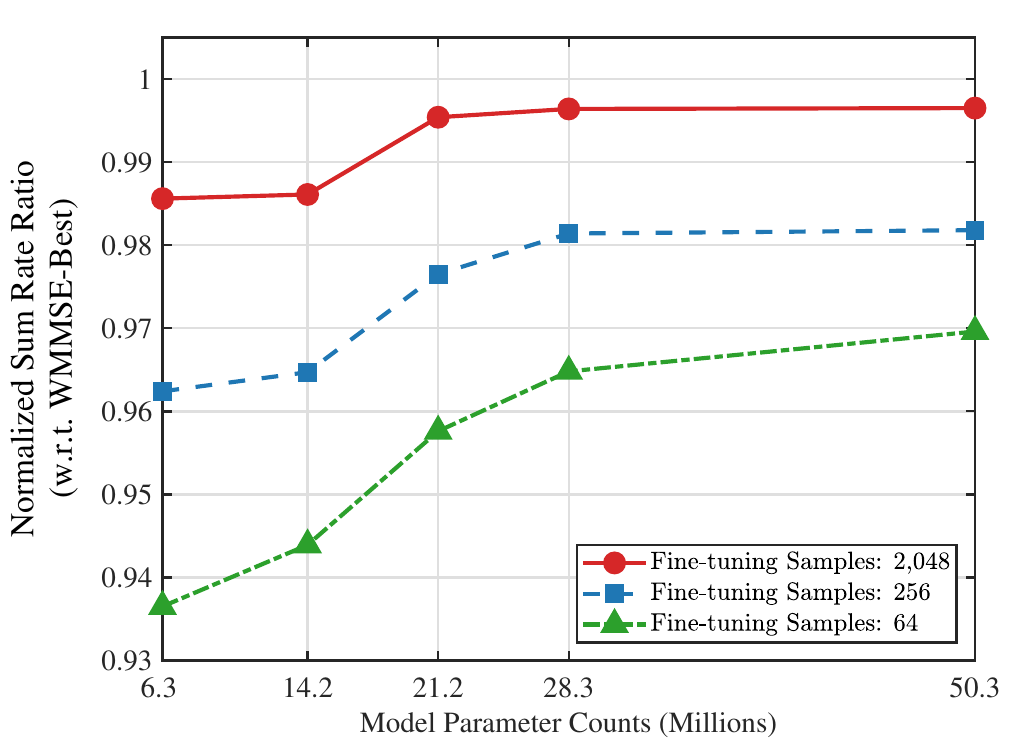}
    \caption{Normalized sum rate performance in the OOD scenario $\mathcal{D}_{18}$ with varying model sizes.}
    \label{fig:scaling_law}
\end{figure}

Fig.~\ref{fig:scaling_law} examines the scalability of the proposed framework by quantifying the relationship between model sizes and performance in the OOD scenario $\mathcal{D}_{18}$. We modulate the complexity of the model by varying the depth ($L \in \{4, 6, 8\}$) and width ($d_{\text{model}} \in$ \{768, 1,024\}), which yields a parameter space ranging from 6.3 million to 50.3 million. Evaluations are conducted under three fine-tuning regimes, i.e., $N_{\text{shot}} \in$ \{64, 256, 2,048\}. As shown in Fig.~\ref{fig:scaling_law}, the normalized sum rate exhibits a monotonic increase as the parameter count grows. Notably, this upward trend persists strongly even in the data-scarce regime ($N_{\text{shot}}=64$) where the metric rises from $0.935$ to $0.97$. This consistent performance gain across different scales demonstrates a clear scaling law within our framework. It indicates that expanding the model capacity inherently enhances its representational power and generalization capabilities. 

\subsection{Ablation Studies}

\begin{figure}[t]
    \centering
    \includegraphics[width=1\linewidth]{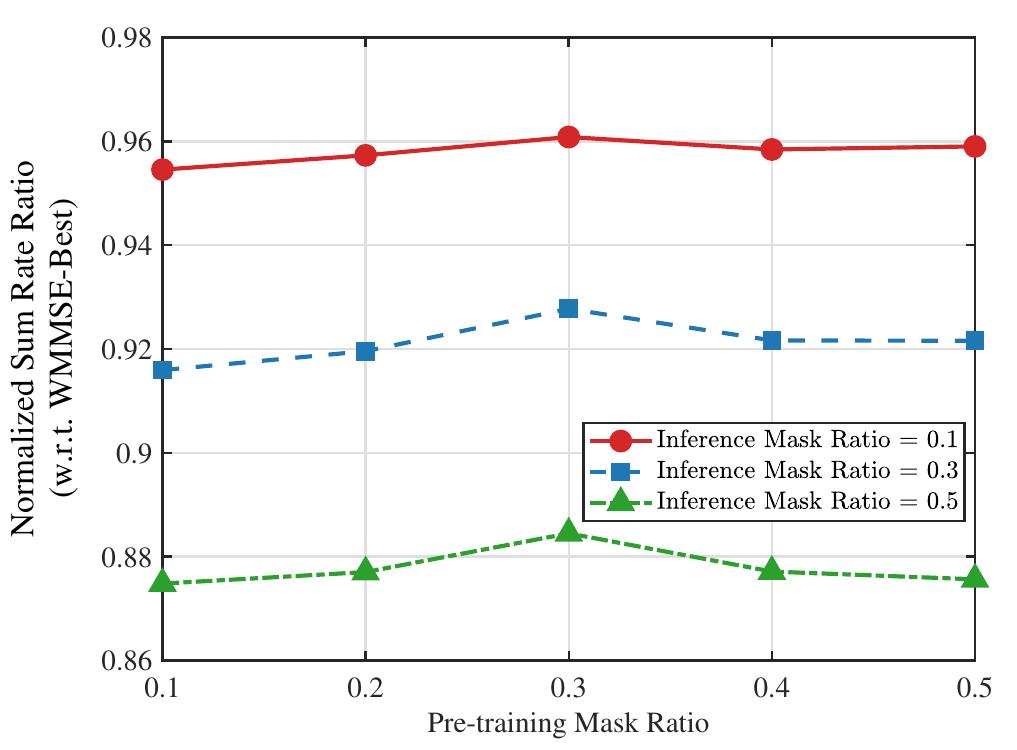}
    \caption{Impact of the pre-training masking ratio $\rho$ on downstream OOD performance in the $\mathcal{D}_{18}$ scenario.} 
    \label{fig:mask_ablation}
\end{figure}

Fig.~\ref{fig:mask_ablation} examines the sensitivity of downstream OOD performance to the pre-training masking ratio $\rho$, utilizing the benchmark scenario $\mathcal{D}_{18}$, i.e., with $K = 50$. To validate these findings under varying stress levels, the normalized sum rate is evaluated across inference mask ratios of $\{0.1, 0.3, 0.5\}$. At lower masking ratios (e.g., $\rho=0.1$), the pre-training task lacks sufficient difficulty because the abundance of visible neighbors allows for trivial local interpolation, which fails to incentivize the capture of global topologies or long-range interference. In contrast, aggressive masking (e.g., $\rho=0.5$) severely fragments the graph structure and destroys the essential context required for reasoning and representation learning. Consequently, $\rho=0.3$ strikes a critical balance, which ensures the task is challenging enough to compel complex structural inference while retaining adequate topological integrity to maximize the robustness of the learned embeddings.

\begin{figure}[t]
    \centering
    \includegraphics[width=1\linewidth]{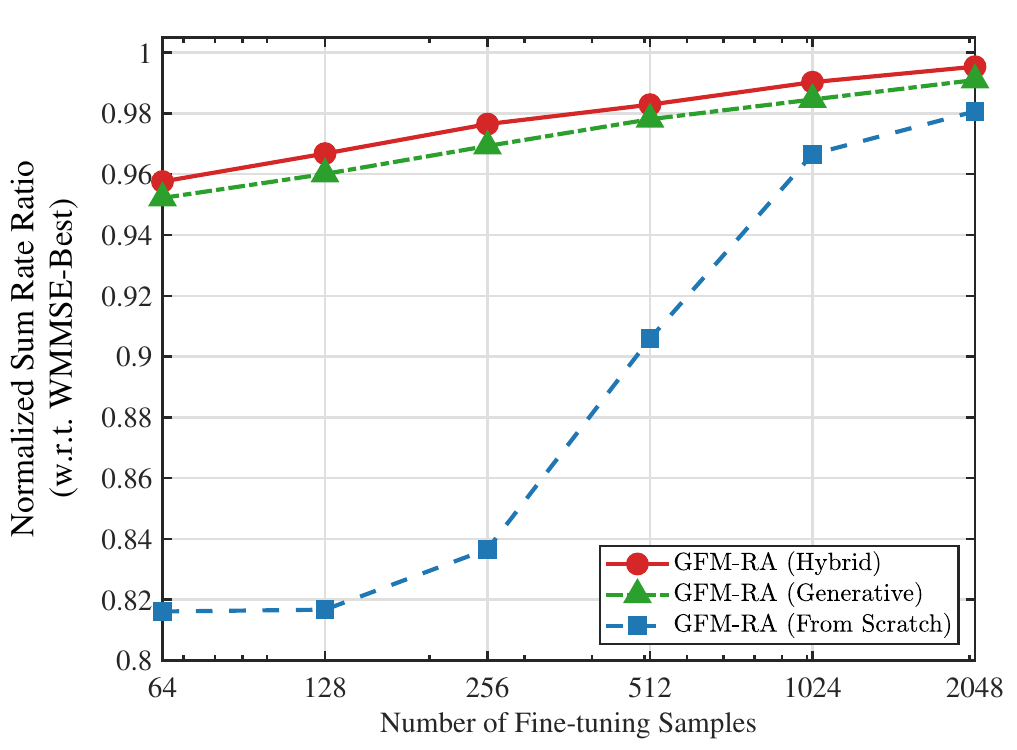}
    \caption{Ablation study in the $\mathcal{D}_{18}$ scenario comparing the proposed hybrid strategy against the edge prediction only variant and the baseline without pre-training.}
    \label{fig:strategy_ablation}
\end{figure}

Fig.~\ref{fig:strategy_ablation} presents an ablation study on the $\mathcal{D}_{18}$ scenario to evaluate the individual contributions of the proposed pre-training objectives. We compare three variants: the uninitialized ``GFM-RA (From Scratch)'', a single-task ``GFM-RA (Generative)'' utilizing only edge prediction, and the complete ``GFM-RA (Hybrid)''.
The results clearly delineate the source of performance gains. ``GFM-RA (From Scratch)'' exhibits lowest performance, reflecting the data-intensive nature of untrained Transformers. Introducing the edge prediction task (``GFM-RA (Generative)'') yields the most substantial leap, boosting the initial sum rate ratio from $0.82$ to $0.95$ at 64 samples. This confirms that generative reconstruction serves as the primary mechanism for capturing local physical interference correlations. Furthermore, ``GFM-RA (Hybrid)'' consistently outperforms the generative-only variant across all sample regimes. This sustained advantage highlights the vital role of the contrastive objective in enforcing global representation robustness against topological perturbations. By synergizing fine-grained local physical reconstruction (generative) with high-level global invariance (contrastive), the hybrid framework achieves optimal sample efficiency and generalization.

\section{Conclusion}

In this paper, we have proposed GFM-RA, a graph foundation model for wireless resource allocation, to establish a universal representation framework capable of seamlessly generalizing across diverse downstream tasks and complex network topologies. By introducing an interference-aware Transformer architecture equipped with a bias projector, we successfully overcome the scalability bottlenecks of conventional message-passing approaches and enable efficient global reasoning over fully connected interference topologies. We further develop a hybrid self-supervised pre-training strategy that combines masked edge prediction and Teacher-Student contrastive learning to distill universal physics from massive unlabeled wireless data. Comprehensive simulation results demonstrate that the proposed framework achieves state-of-the-art performance and exhibits superior scalability, where increased model capacity translates to measurable performance gains.  Moreover, by leveraging its pre-trained robust representations, the model displays exceptional sample efficiency, enabling rapid adaptation to diverse downstream optimization objectives and OOD scenarios with strictly limited samples. This work validates the viability of the pre-training and fine-tuning paradigm in wireless resource allocation and paves the way for developing general-purpose foundation models for 6G intelligent networks.

\small
\bibliographystyle{IEEEtran}
\bibliography{journal_dc.bib}

\end{document}